# A Novel Community Detection Based Genetic Algorithm for Feature Selection

**Mehrdad Rostami[1], Kamal Berahmand[2], Saman Forouzandeh[3]**

*Department of Computer Engineering, University of Kurdistan, Sanandaj, Iran[1]*
*Department of Science and Engineering, Queensland University of Technology, Brisbane, Australia[2]*
*Department of Computer Engineering University of Applied Science and Technology, Center of Tehran Municipality ICT org.Tehran, Iran[3]*
*M.rostami@eng.uok.ac.ir[1], kamal.berahmand@hdr.qut.edu.au[2], Saman.forouzandeh@gmail.com[3]*

**Abstract**
The selection of features is an essential data preprocessing stage in data mining. The core principle of feature selection seems to be to pick a subset of possible features by excluding features with almost no predictive information as well as highly associated redundant features. In the past several years, a variety of meta-heuristic methods were introduced to eliminate redundant and irrelevant features as much as possible from high-dimensional datasets. Among the main disadvantages of present meta-heuristic based approaches is that they are often neglecting the correlation between a set of selected features. In this article, for the purpose of feature selection, the authors propose a genetic algorithm based on community detection, which functions in three steps. The feature similarities are calculated in the first step. The features are classified by community detection algorithms into clusters throughout the second step. In the third step, features are picked by a genetic algorithm with a new community-based repair operation. Nine benchmark classification problems were analyzed in terms of the performance of the presented approach. Also, the authors have compared the efficiency of the proposed approach with the findings from four available algorithms for feature selection. The findings indicate that the new approach continuously yields improved classification accuracy.

## 1. Introduction

Data sets have evolved significantly in recent years with developments in science and technology and now involve numerous features. Methods of pattern detection are therefore engaged in samples with thousands of features. Consequently, reducing their dimensionality is essential for traceability of data sets [1, 2]. High-dimensional vectors impose significant computational costs and also the risk of overfitting. Generally, a minimum of 10×D×C training examples is necessary for a classification problem with D dimensions and C classes [3]. Whenever the needed number of training examples cannot be provided, reducing features decreases the size of the needed training examples and hence increases the overall yield shape of the classification algorithm. In the previous years, two methods for dimensional reduction were presented: feature selection and feature extraction. Feature selection (FS) seeks for a relevant subset of existing features, while features are designed for a new space of lower dimensionality in the feature extraction method. Both methods for the reduction of dimensionality are designed to improve learning efficiency, minimize computational complexity, develop more generalizable models, and reduce needed storage [4-6].

Techniques of optimization based on the population including ant colony optimization (ACO) [7], genetic algorithm (GA) [8], simulated annealing (SA) [9], taboo search (TS) [10], and particle swarm optimization (PSO) [11] were recently used in feature selection. In fact, hybrid search strategies have used that merge the wrapper and filter approaches. In[12], the suggestion was made for the use of a hybrid filter wrapper subset selection algorithm based on the PSO for the classification of support vector machines (SVM). In addition, some existing techniques take into account the connection of features in their search strategies. For instance, in [13], an enhanced genetic algorithm was proposed for the optimum selection of a feature subset from a multi-character set. This approach separates the chromosome into many classifications for local management. Various mutation and crossover operators are then used on mentioned categories to eliminate invalid chromosomes. In recent decades, many Evolutionary algorithms-based algorithms such as Genetic Algorithm (GA), Particle Swarm Optimization (PSO), Ant Colony Optimization (ACO), and Artificial Bee Colony (ABC) have been employed to feature section. Among the SI-based algorithm, Genetic has been efficiently utilized in the feature selection problem to redact of high-dimensional dataset. One of the disadvantages of this method is that it does not consider the connections among the features when selecting the final features. As a result, the probability of selecting a subset with redundancy will increase. To overcome these drawbacks, the present paper introduces a community-based genetic algorithm for the selection of features named CGAFS. A community detection method is used in the proposed approach for dividing features into various classes. Hence a new mutation operator named "repair operations" is introduced to fix the chromosome by utilizing predetermined feature clusters. A newly produced offspring shall be repaired to eliminate related features in the offspring. In comparison to the previous genetic algorithm-based feature selection that apply filters and wrappers models in the order, the community detection technique is integrated into the GA-based wrapper model in a structural manner. Furthermore, the cluster number and the optimum size of the subset could also be calculated automatically. The proposed methods have several novelties compared to the well-known and state-of-the-art GA-based feature selection methods:

- ✓ The proposed method uses a novel community detection-based algorithm to identify the feature clusters to group similar features. Grouping similar features prevent the proposed method to select redundant features. Unlike the other clustering methods such as k-means [14] and fuzzy c-means [15], the proposed clustering method identifies the number of clusters automatically, and there is no longer a need to determine the number of clusters in advance.
- ✓ The proposed method uses community detection-based repair operation which considers both local and global structure of the graph in computing similarity values. In other words, it takes into account implicit and explicit similarities between features, while the other feature selection methods only take into account the direct similarities between features.
- ✓ The number of final selected features imposes another challenge on feature selection methods. In other words, the number of relevant features is unknown; thus, the optimal number of selected features is not known either. In this method, unlike many previous works, the optimal number of selected features is determined automatically based on the overall structure of the original features and their inner similarities.
- ✓ The proposed method groups similar features into the clusters and then applies a multi-objective fitness function to assign an importance value to each feature subset. In the proposed multi-objective fitness function, two objectives of feature relevance and feature redundancy are considered, simultaneously. Unlike the other multi-objective methods that identify a set of non-dominated solutions in an iterative process [16, 17], the proposed method finds the near-optimal solution in a reasonable time.

The rest of the present article is structured as the following: Section 2 analyses research on the selection of features; in Section 3, the proposed selection algorithm is presented; in Section 4, the comparison of the proposed algorithm other feature selection algorithms is discussed. Ultimately, in Section 5, the authors summarize the present study.

## 2. Related Work

For several practical applications, including text processing, face recognition, image retrieval, medical diagnosis, and bioinformatics, feature selection was developed as a central procedure [18-20]. Feature selection was a promising area of research and development for statistical pattern detection, data mining and machine learning since the 1970s, and many efforts have been made to evaluate the methods of feature selection, which may be divided into four groups, namely, filters, wrappers, hybrids and embedded depending on the evaluation process [21-24]. Whenever a procedure performs a feature selection independently of any learning algorithm (e.g., an entirely independent preprocessor), afterward it is included in the filter method classification. The statistical analysis is required for the filter approach of the feature set that can only be used to solve the feature selection problem without using a learning model. Conversely, a predetermined learning algorithm is used by the wrapper approach to identify the quality of the selected subsets. However, wrappers can yield stronger results; they are costly to operate and can disintegrate with too many features. The hybrid approach combines the filter and wrapper technique and seeks to incorporate the filter and wrapper methods. Ultimately, the embedded techniques take advantage of the selection of features in the learning process as well as are highly comparable to a certain learning model.

Depending on the availability of training data class labels, future selection algorithms could also be classified into two parts: supervised feature selection and unsupervised feature selection [25, 26]. The supervised feature selection is employed in the case that class labels of the data are obtainable, differently the unsupervised feature selection seems to be suitable. In general, the supervised feature selection generates better and more efficiency, primarily because of the use of class labels.

From another view, filter methods are classified into ranking-based and subset selection-based (SSB) methods. Ranking-based methods first assign a relevance value to each feature using a univariate or a multivariate criterion, and then sort the features and select those of the top high scores. Although the ranking-based methods require low computational resources, all these methods consider only the relevancy of the features and neglect the redundancy with others. Identifying a set of optimal feature subset that results in building a learning model with maximum accuracy is an NP-hard problem. To overcome this issue, the subset selection-based methods seek to find a near-optimal feature set by applying some heuristic or meta-heuristic methods. For example, Relevance redundancy feature selection (RRFS) [27], MIFS [28], Normalized mutual information feature selection (NMIFS) [29], MIFS-U [30], MIFS-ND [31], JMIM [32], OSFMI and MRDC [33] use sequential forward or backward selection as a type of greedy search strategy, and thus they easily trap into a local optimum.

The search space includes all feasible feature subsets to discover the best feature subset, indicating that the search space is as the following:

$$\sum_{s=0}^{n}\binom{n}{s}=\binom{n}{0}+\binom{n}{1}+\ldots+\binom{n}{n}=2^{n} \quad (1)$$

Where $n$ (quantity of original features) is the dimensionality and s is the size of the current subset of features. Thus, the problem to discover the ideal feature subset seems to be NP-hard. Because the analysis of the whole feature subsets is costly in a computational manner, time-consuming and also inefficient even in small sizes, solutions are required that are computationally efficient and that provide a reasonable tradeoff among time-space cost and strength of the solution. Most feature selection algorithms also include random or heuristic search techniques to minimize the computation period [34-36].

One approach to solving complex optimization and NP-Hard problems is meta-heuristics algorithms. Meta-heuristic algorithms are approximate approaches that can find satisfactory solutions over an acceptable time instead of finding the optimal solution [37]. These algorithms are one of the categories of approximate optimization algorithms that have s strategies to escape from local optima and can be used in a wide range of optimization problems.

Many feature selection methods use meta-heuristics to avoid increasing computational complexity in the high dimensional dataset. These algorithms use primitive mechanisms and operations to solve an optimization problem and search for the optimal solution over several iterations [38]. These algorithms often start with a population containing random solutions and try to improve the optimality of these solutions during each iteration step. At the beginning of most of the meta-heuristic algorithms, a number of initial solutions are randomly generated, and then a fitness function is utilized to calculate the optimality of the individual solutions of the generated population. If none of the termination criteria are met, production new generation will begin. This cycle is repeated until one of the termination criteria is met [39, 40].

Meta-heuristic approaches can be classified into two categories: Evolutionary Algorithms (EA) and Swarm Intelligence (SI) [37]. An EA uses mechanisms inspired by biological evolution, such as reproduction, mutation, recombination, and selection. Candidate solutions to the optimization problem play the role of individuals in a population, and the fitness function determines the quality of these solutions. After repetitions of the evolutionary algorithm, the initial population evolves and moves toward global optimization[41]. On the other, SI algorithms usually consist of a simple population of artificial agents locally with the environment. This concept is usually inspired by nature, and each agent performs an easy job, but local interactions and partly random interactions between these agents lead to the emergence of "intelligent" global behavior, which is unknown to individual agents[42].

## 3. Proposed Method

For real-world datasets, there are a vast number of irrelevant and redundant features, which may significantly degrade the performance of the model learned and the learning speed of the models. Feature selection is an essential step in data preprocessing in data mining to remove irrelevant and redundant features of a given dataset. Many technologies can easily eliminate irrelevant features from the other feature subset selection methods, but do not handle redundant features. Many often only eliminate redundant features. With the redundant features, the presented algorithm will remove the irrelevant.

The authors consider a hybridized method based on a combination of a community detection approach and the genetic algorithm, in the context of the hybrid approaches to the feature selection problem. Genetic algorithms are methods of optimization focused on the natural selection process. John Holland initially introduced GAs to describe the adaptation mechanisms of the natural systems and to develop new artificial structures on identical principles. This imitates the natural selection method and begins with artificial individuals (represented by a 'chromosome' population). GA attempts to improve the fitters using genetic operators (e.g., crossover and mutation). In addition, it seeks to produce chromosomes in a certain quantitative measure, which are stronger compared to their parents. Hence, GA has recently been widely used as a tool for data mining feature selection.

In theory, it was shown that genetic algorithms could randomly seek the optimal solution for a problem. Simple genetic algorithms, however, have some shortcomings such as premature convergence, poor ability of fine-tuning near local optimum points in applications. On the other side, certain other techniques of optimizing, including the steepest descent method, simulated annealing, and hill-climbing generally include strong local searchability. Moreover, some heuristic algorithms have a strong performance with issue-specific information. Furthermore, some hybrid GAs for feature selection was established by incorporating the optimization methods or heuristic algorithms, as mentioned above, to improve the fine-tuning capabilities and performance of simple GAs. In the present study, the authors suggest a new genetic algorithm of clustering for feature selection issues, in which the connection and repair of this feature are used for the selection of candidate features.

Application of the hybrid genetic algorithm for the selection of features typically involves chromosome encoding schemes, fitness function estimation, fitter chromosome selection, genetic crossover and mutation operations, and stoppage criterion. The suggested approach provides a candidate solution to the problem of subset selection in the chromosome population. A chromosome is encoded with binary digit series that "1" means "selected" and "0" means "unselected." Every digit (or gene) correlates to a feature so that the chromosome gene length is equivalent to the total of input features available. The methods for genetic operations are as follows. Initially, the design proposed in the present article uses the roulette wheels' selection process. Next, an adaptive crossover approach is applied. The single-point crossover operator is utilized where the overall number of features in a specified dataset is less than 20; whereas the overall number of functions is greater than 20, double-point crossover procedures are used.

The main CDGAFS steps are summarized in Fig. 1. In addition, in its corresponding subscription, every stage of the CDGAFS is defined.

**Step1: Measure the fisher score of features:**

For measuring the discriminatory power of the features, the discrimination ability of the feature $F_i$ is measured by applying the Fisher score as the following:

$$Score_i = \frac{\sum_{k=1}^{C} n_i (\bar{x}_i^k - \bar{x}_i)^2}{\sum_{k=1}^{C} n_i (\sigma_i^k)^2} \qquad (2)$$

Where, $C$ implies the number of classes of the dataset; $n_i$ is referred to as the number of samples in class i, $\bar{x}_i$ indicates the mean of all the patterns according to the feature $F_i$, as well as $\bar{x}_i^k$ and $\sigma_i^k$ imply mean and variance of class $k$ corresponding to the feature $F_i$. A larger $Score_i$ value shows that the feature $F_i$ possesses a higher discriminative capability. In most instances, fisher score values of features are near each other. In order to conquer this situation, a non-linear normalization approach named softmax scaling has been applied for scaling the edge weight into the range [0 1] as the following:

$$\widehat{Score}_i = \frac{1}{1 + \exp\left(-\frac{Score_i - \overline{Score}}{\sigma}\right)} \tag{3}$$

Where $Score_i$ indicates the fisher score of the feature $F_i$, $\overline{Score}$ and $\sigma$ imply the variance and mean of all of the fisher score values, respectively, as well as $\widehat{Score}_i$ shows normalized fisher score value of the feature $F_i$.

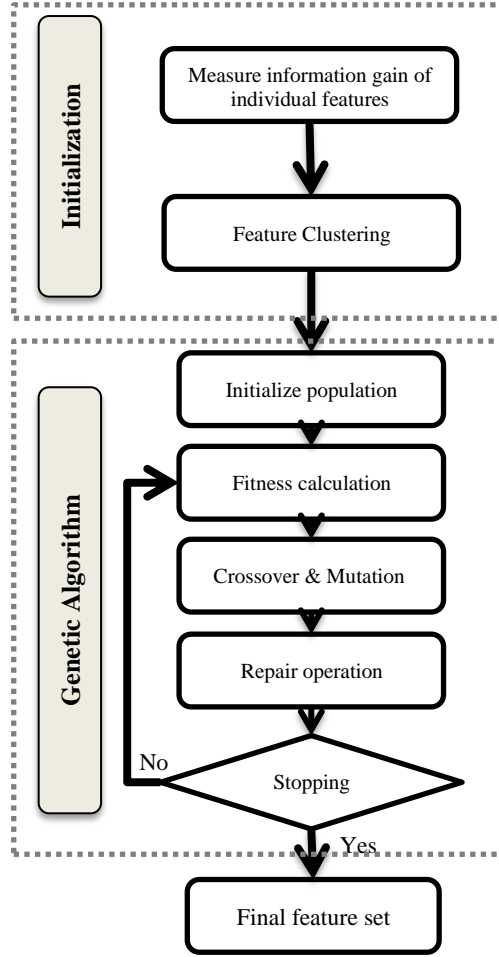

**Fig 1.** Flowchart of the proposed method

## Step2: Irrelevant Feature Removal:

Many databases provide a wide collection of features. Thus, it is not efficient to manipulate such datasets for the classification since certain features are not effective for classification. In order to tackle this issue, the authors reduced the dataset to 100 features by applying a fish score measurement technique for the first time.

## Step3: Feature clustering:

The central point is the clustering of the original features into many clusters based on the similarities among features. The identical cluster's features are also the same. The majority of known clustering techniques have certain limitations. First, showing the desired number of clusters, the parameter k, should be determined in the following. Identifying the appropriate quantity of clusters usually involves exhaustive trial and error. Second, the data distribution in a cluster is a major factor, and the variance of similarity estimation of the underlying cluster is not considered in current approaches. Thirdly, all features in a cluster lead to the subsequent derived function. To address these problems in this article, a graph-based community detection approach is applied to classify the features. The authors employed the community detection algorithm of Louvain to define the feature clusters. This algorithm determines the communities in the graph by optimizing a modularity function that is an efficient and simple way for identifying communities in broad networks.

Generally, the feature space must be defined by a completely connected undirected graph for using the Louvain community detection algorithm. Therefore, the authors try to model the problem of feature selection by employing a theoretical graph representation. For this purpose, the mapping of the feature set into its equivalent graph $G = (F, E, w_F)$ was done, where $F = \{F_1, F_2, \ldots, F_n\}$ implies a set of original features, $E = \{(F_i, F_j): F_i, F_j \in F\}$ are the edges of the graph and $w_{ij}$ is referred to as the similarity among two features $F_i$ and $F_j$ which were connected by the edge $(F_i, F_j)$. In the present article, the Pearson correlation coefficient measure has been applied to calculate the similarity value among various features of a provided training set. The relationship between the two features $F_i$ and $F_j$ is defined as the following:

$$w_{ij} = \left| \frac{\sum_p (x_i - \overline{x_i})(x_j - \overline{x_j})}{\sqrt{\sum_p (x_i - \overline{x_i})^2} \sqrt{\sum_p (x_j - \overline{x_j})^2}} \right| \quad (4)$$

Where $x_i$ and $x_j$ imply the vectors of features $F_i$ and $F_j$ in a respective manner. The variables $\overline{x_i}$ and $\overline{x_j}$ denote the mean values of vectors $x_i$ and $x_j$, averaged over $p$ samples. Obviously, the similarity value among a couple of completely similar features will be 1, and on the other hand, this value will be equal to 0 for entirely dissimilar features. Similar to fisher score values, all similarity values are normalized by the softmax scaling method.

After the generation of feature graphs, the initial nodes are divided into a number of clusters in such a way that the members of each cluster have the maximum similarity levels with respect to each other. Most of the existing feature clustering methods suffer from one or more of the following shortcomings[1]:

- the need to specify the number of clusters before performing feature clustering;
- the distribution of features in a cluster, which is one of the most important criteria in feature clustering, is not considered;
- all features are considered equally, while certain influential features should have a greater impact on the clustering process

To deal with these issues, an iterative search algorithm for community detection (ISCD) [43] is applied to cluster the features in this study. The ISCD algorithm can quickly detect communities, even in large graphs, due to the linear computational complexity. As such, it is efficient for feature clustering of high-dimensional data.

### Step 4: Initialize Population:

A population set of chromosomes is produced in this step in a random manner. The number of original features n is equal to each chromosome length. Each chromosome gene is given a value of 1 or 0. When a feature is chosen, the respective gene in the chromosome is set to 1; otherwise, the gene value is set to 0. It is noteworthy that the total number of selected features in each chromosome must be $k \times \omega$, where $k$ implies the number of clusters, and $\omega$ is a user-specified parameter controlling the size of the final feature subset.

### Step 5: Calculate Fitness values:

After creating the initial population, the fitness function for all chromosomes must be calculated. For this purpose, in this proposed method, a novel multi objective fitness function is introduced. In this fitness function, a combination of classification accuracy in the KNN classification algorithm and the sum of similarities between the selected features is used. The fit of the $FS^k$ feature subset in the iteration $t$ denoted by $J(FS^k(t))$ is measured by Equation (5).

$$J\left(FS^{k}(t)\right) = \frac{CA\left(FS^{k}(t)\right)}{\frac{2}{\left|FS^{k}(t)\right|*\left(\left|FS^{k}(t)\right|-1\right)} \sum_{F_i,F_j \in FS^k} Sim(F_i,F_j)} \quad (5)$$

Where, $CA(FS^k(t))$ indicates the classification accuracy for the selected feature subset $FS^k(t)$ on the KNN classifier, $|FS^k(t)|$ represents the subset size the selected features $FS^k(t)$ and $Sim(F_i, F_j)$ indicates the similarity between the attribute $F_i$ and $F_j$. As can be seen in this Equation, in calculating the suitability of each subset, the classification accuracy for that subset and the total similarity between the features selected in that subset are considered simultaneously. Consequently, a higher set of features is allocated to the feature's subset possessing the most relevance to the objective class and the least redundancy.

### Step 6: Perform Crossover & Mutation operation:

New chromosomes are produced by crossover and mutation operators [38]. The single point crossover among the selected chromosomes has been used in this research to produce new populations. In addition, a single parent chromosome may be flipped by randomly flipping one or more bits to create a child. That chromosome gene follows the predefined probability of mutation, whether or not it chooses to be mutated.

### Step 7: Perform Repair Operation:

The proposed technique suggests a repair operation on an offspring among all freshly created chromosome to re-adjust the number of features selected from every group. If the number of selected features in one of the clusters is less than $\omega$, one feature is randomly selected, and the corresponding gene is adjusted to be 1. Moreover, where more than one feature has been selected, one of them is randomly retained, and the other is eliminated from the chromosome. The repair process includes the unique and general characteristics of a certain dataset for the offspring generated by the fitter. Two steps are regarded for the repair in CDGAFS: (i) the check of the number of features in each cluster; and (ii) the enhancement of the offspring. It is noteworthy that only once will the first stage be done. The details of the repair procedure are shown in Figure 2.

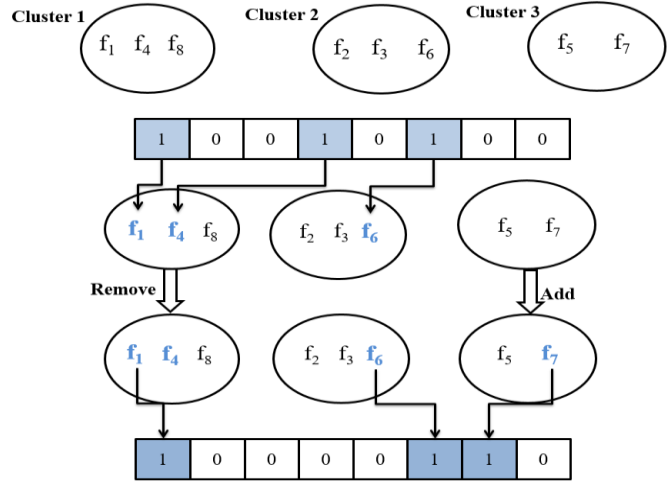

**Fig 2.** Details of repair operation of the proposed method

As an example, assume that number of features and $\omega$ are set to 3 and 1. The repair operation stages described further as the following: (a) newly generated one offspring, (b) features are divided by developing 3 subgroups according to the comparison of features from clusters $C_1$, $C_2$ and $C_3$, respectively, (c) re-arrangement of features of these subgroups is done, (d) produce repaired offspring. The aim of repair operation is to direct the search procedure in order that the recently produced offspring is capable of being adjusted by the less associated features. Therefore, the presented

CDGAFS discovers non-redundant features from original features; the meaning of this process is that the resultant features do not correlate with each other.

**Step 8: Stopping Criterion:**

In the case that the number of iterations is higher than the maximum allowable iteration, continue; otherwise, take a step in the fitness calculation.

**Step 9: Final Subset Selection:**

Eventually, according to its fitness value, the strongest chromosome of the last generation indicates the optimal subset of features for a specific dataset.

Algorithm 1 shows the pseudo-code of the proposed method.

| **Algorithm 1: Community Detection-based Genetic Algorithm for Feature selection (CDGAFS)** | |
|---|---|
| **Input** | $D_T$: Training dataset |
|  | $\omega$: Selected feature from each community |
| **Output** | $F' = \{f'_1, \dots, f'_m\}$ |
| 1: | **Begin algorithm** |
| 2: | Measure the fisher score of features |
| 3: | Remove the irrelevant features |
| 4: | Calculate feature similarities |
| 5: | $\{C_1, C_2, \dots, C_k\}$ = Apply community detection algorithm to create feature clusters |
| 6: | Generate the initial population of chromosomes |
| 7: | **Do** |
| 8: |     Calculate Fitness values |
| 9: |     Perform Crossover & Mutation operation |
| 10: |     Perform Repair Operation |
| 11: | **While until (The Stopping Criterion is met)** |
| 12: | Select the best chromosome |
| 13: | Report $F'$ as final feature set |
| 14: | **End algorithm** |

## 4. Experiments and discussion

Many tests were carried out for both the classification accuracy and the number of selected features to assess the proposed approach. The findings have been discussed in this section. The experiments were conducted on a 3.58 GHz CPU and 8 GB RAM machine.

In these experiments, one feature selection method was chosen and evaluated in the experimental result for comparing the efficiency of various techniques of feature selection based on each SI-based algorithm. For a fair evaluation, all of the methods examined in this section were selected from among wrapper-based methods. These wrapper-based methods include PSO-based [44], ACO-based [45], and ABC-based [46]

### 4.1 Datasets and preprocessing

The efficiency of CDGAFS was provided in this regard on six popular benchmark classification datasets, i.e., SpamBase, Sonar, Arrhythmia, Madelon, Isolet, and Colon. Several of these datasets include characteristics with missing values so that each missing value was substituted with the average of the data present on the corresponding feature to cope with these values in the tests. Furthermore, in many practical situations, a designer is faced with features; the values of these features are in various ranges. The features associated with a broad range of values thus dominate those related to small range values. A non-linear normalization approach named softmax scaling is applied to measure the datasets to solve this problem.

After the normalization process, each dataset was randomly partitioned into three subsets, such as validation set, training set, and testing set. The distribution of the number of instances and features of these datasets is presented in Table 1.

Table 1: Characteristics of the used medical datasets

| Dataset | Features | Classes | Patterns |
|---|---|---|---|
| SpamBase | 57 | 2 | 4601 |
| Sonar | 60 | 2 | 208 |
| Arrhythmia | 279 | 16 | 351 |
| Madelon | 500 | 2 | 4400 |
| Isolet | 617 | 26 | 1559 |
| Colon | 2000 | 2 | 62 |

## 4.2 User-specified parameters

Certain user-specified parameters are available to be defined for CDGAFS to work appropriately for the feature selection task. It is noteworthy that these parameters are non-unique to the suggested process, however, are needed for any GA-based feature selection algorithm. After some preliminary implementation, the parameters were selected and are not meant to be optimum. Table 2 demonstrates the common parameters for all datasets.

Table 2: User-specified parameters of the GA-based method

| Parameter | Values |
|---|---|
| Crossover rate | 0.8 |
| Mutation rate | 0.05 |
| Population Size | 100 |
| Iteration number | 100 |

## 4.3 The utilized Classifier

For assessing the generalizability of the presented approaches in various classifiers, in these tests, 3 classifiers, such as Support Vector Machine (SVM), AdaBoost (AB), and K Nearest Neighbors (KNN) are utilized.

In pattern recognition, the KNN classifier is a non-parametric approach presented for regression and classification. In both cases, the input contains the nearest examples of training in the feature space. Support vector machine SVM is among Vapnik's supervised learning algorithms. The purpose of SVM is the maximization of the margin among data samples, and excellent performance for classification and regression problems has been shown recently. AdaBoost (AB) ("Adaptive Boosting") is a meta-algorithm for machine learning formulated by Yoav Freund and Robert Schapire. The AdaBoost classifier is a meta-estimator starting with the fitting of a classifier and fitting of additional copies on the identical dataset, afterward the weights of improperly grouped examples are modified to concentrate on severe cases more in subsequent classifiers. Weka (Waikato Environment for knowledge analysis) is the experimental workbench [47], a set of data mining methods. In the present study, KNN, AdaBoostM1, and SMO as the WEKA implementation of KNN, AB, and SVM have been applied.

## 4.4 Results

In these experiments, the feature subset size and classification accuracy are used as the performance evaluation criteria. In the experiments, first, the comparison of the performances of different wrapper SI-based feature selection approaches is done with various classifiers. Tables 3 presents the mean classification accuracy (%) over 10 independent runs of the various SI-based wrapper feature selection techniques by employing AB, NB, and SVM classifiers. Each entry of Tables 4 implies the mean value and standard deviation (given in parenthesis) of 10 independent runs. The optimal result is demonstrated in an underlined and boldface, and the second-best is in boldface. Table 4 shows that, in the majority of cases, the performance of the proposed CDGAFS approach is better compared to the other evolutionary-based feature selection method. For instance, in the SpamBase dataset on the KNN classifier, the proposed method obtained a 93.99 % classification accuracy. In contrast, for PSO, ACO, and ABC methods, these values were reported 92.54 %, 91.81%, and 90.35 %, correspondingly.

**Table 3:** Average classification accuracy rate and as standard deviation (shown in parenthesis) over ten runs of the evolutionary-based feature selection methods using KNN, SVM, and AdaBoost classifier. The best result is indicated in boldface and underlined, and the second-best is in boldface.

| Dataset | Method | Classifier | | |
| --- | --- | --- | --- | --- |
| | | KNN | SVM | AdaBoost |
| SpamBase | PSO | **92.54 (2.83)** | **92.35 (1.52)** | **92.43 (2.25)** |
| | ACO | 91.81 (1.82) | 89.51 (2.81) | 90.89 (2.21) |
| | ABC | 90.35 (2.39) | 89.22 (1.93) | 91.30 (3.33) |
| | CDGAFS | <u>**93.99 (2.76)**</u> | <u>**93.68 (1.73)**</u> | <u>**93.27 (2.82)**</u> |
| Sonar | PSO | **88.18 (2.43)** | **87.81 (2.29)** | **86.93 (3.32)** |
| | ACO | 88.76 (2.32) | 87.36 (3.32) | 85.82 (1.48) |
| | ABC | 87.23 (1.13) | 87.17 (2.81) | 86.74 (1.78) |
| | CDGAFS | <u>**88.73 (3.76)**</u> | <u>**88.34 (2.19)**</u> | <u>**87.13 (2.71)**</u> |
| Arrhythmia | PSO | **86.15 (2.82)** | 86.01 (2.61) | **85.91 (2.82)** |
| | ACO | 84.13 (2.12) | **86.27 (2.62)** | 85.72 (3.94) |
| | ABC | 85.83 (2.73) | 85.71 (1.75) | 84.32 (1.39) |
| | SSA | <u>**87.21 (2.37)**</u> | <u>**87.38 (2.02)**</u> | <u>**86.98 (2.59)**</u> |
| Madelon | PSO | 86.46 (3.14) | 86.65 (2.47) | 86.12 (1.81) |
| | ACO | 86.19 (2.20) | 85.91 (1.32) | **86.34 (2.11)** |
| | ABC | **87.55 (2.13)** | **87.19 (1.81)** | 86.12 (2.33) |
| | SSA | <u>**87.88 (1.55)**</u> | <u>**87.82 (1.64)**</u> | <u>**86.79 (1.62)**</u> |
| Isolet | PSO | **85.63 (1.39)** | 85.39 (1.62) | **85.42 (2.32)** |
| | ACO | 85.26 (1.58) | **85.90 (1.81)** | 85.41 (2.39) |
| | ABC | 84.38 (2.81) | 84.95 (2.16) | 84.84 (1.48) |
| | SSA | <u>**86.11 (2.44)**</u> | <u>**86.01 (2.65)**</u> | <u>**85.39 (2.62)**</u> |
| Colon | PSO | <u>**96.41 (2.82)**</u> | <u>**96.19 (2.16)**</u> | <u>**96.32 (1.31)**</u> |
| | ACO | 94.43 (1.71) | 95.73 (1.19) | 96.12 (1.82) |
| | ABC | 93.04 (2.56) | 92.61 (3.61) | 92.49 (3.45) |
| | SSA | **95.41 (2.15)** | **95.82 (2.65)** | **95.36 (2.38)** |

Moreover, Figures 3 to 5 show the mean classification accuracy over all datasets on the KNN, SVM, and AdaBoost classifiers, respectively. As can be seen in these figures, on all classifiers, the suggested approach had the highest average classification accuracy. The findings presented in Figure 3 indicate that the presented technique obtained 89.89 % mean classification accuracy and obtained the first rank with a 0.66% margin in comparison with the PSO-based approach, which achieved the second-best average classification accuracy. Moreover, results presented in Figure 4 show the discrepancies among the achieved classification accuracy of the suggested technique and the second-best ones (PSO-based) and third-best ones (ACO-based) on SVM classifier were reported 0.77 (i.e., 89.84-89.07) and 1.39 (89.84-88.45) percent. Furthermore, based on the result of Figure 5, on the AB classifier, the proposed CDGAFS method gained the first rank with an average classification accuracy of 89.15 %, and the ACO-based and PSO-based feature selection techniques were ranked second and third with an average classification accuracy of 88.86 % and 88.38 %, respectively.

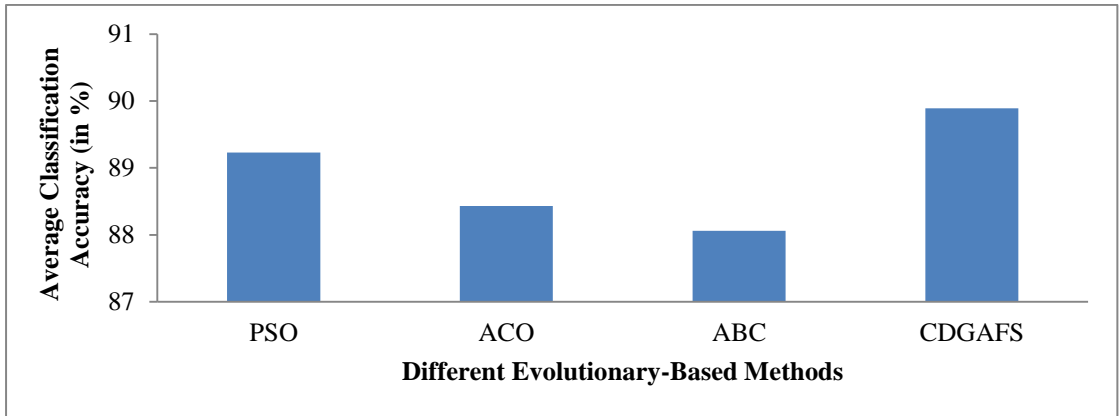

**Figure 3:** Average classification accuracy over all datasets on the KNN classifier.

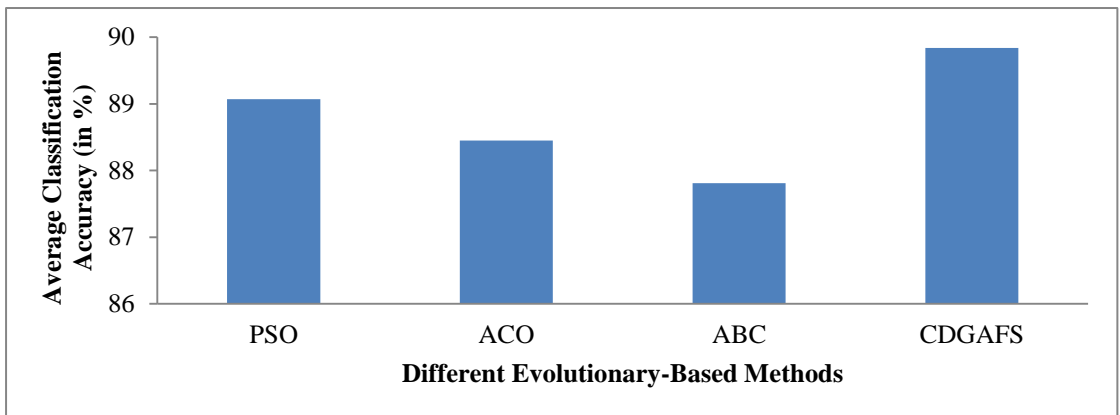

**Figure 4:** Average classification accuracy over all datasets on the SVM classifier

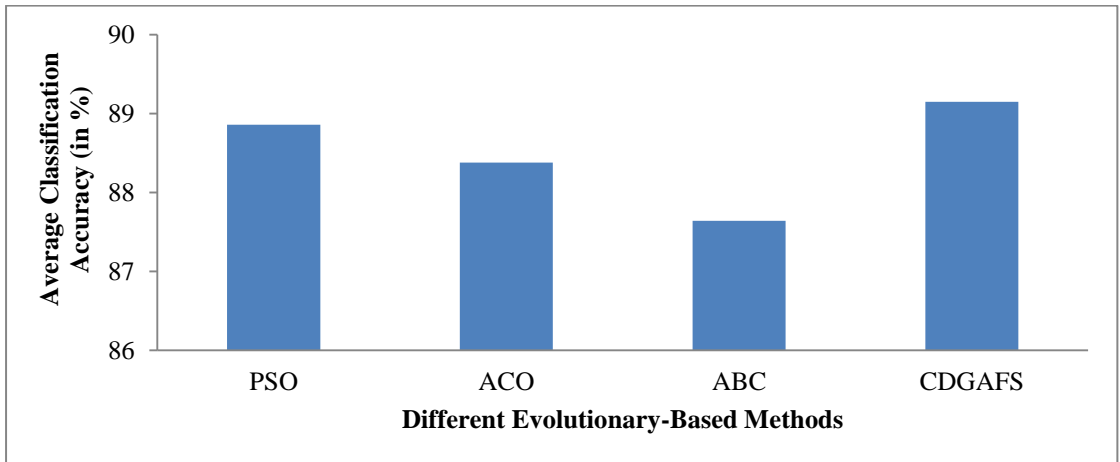

**Figure 5:** Average classification accuracy over all datasets on the AdaBoost classifier

Table 4 records the number of selected features of the four wrappers evolutionary-based feature selection approaches for each dataset. It is evident that in a general manner, all the four approaches obtain a considerable decrease of dimensionality by choosing a small part of the original features. Among various methods, in SpamBase, Sonar, Arrhythmia, Isolet datasets, the proposed technique shows the best performance compared to the other evolutionary-based approaches, selecting only 14.33, 10.52, 7.06, and 21.92 %, respectively. Moreover, in the Madelon and Colon datasets, the PSO-base method selected an average of 14.87 and 0.64 features, respectively. In Madelon and Colon datasets, the proposed feature selection method was ranked second with a mean classification accuracy of 15.01 % and 0.65 %, respectively.

**Table 4:** Average number of selected features of the different wrapper evolutionary-based methods. (Minimum number of selected features is indicated in boldface and underlined and the second best is in boldface)

| Dataset | Number of the original feature | Method | Number of selected features | The ratio of the selected features to the original features (in %) |
|---|---|---|---|---|
| SpamBase | 57 | PSO | 8.92 | 15.65 |
| | | ACO | **8.87** | **15.56** |
| | | ABC | 8.91 | 15.63 |
| | | CDGAFS | <u>**8.17**</u> | <u>**14.33**</u> |
| Sonar | 60 | PSO | 7.31 | 12.18 |
| | | ACO | **7.13** | **11.88** |
| | | ABC | 7.83 | 13.05 |
| | | CDGAFS | <u>**6.31**</u> | <u>**10.52**</u> |
| Arrhythmia | 279 | PSO | 20.12 | 7.21 |
| | | ACO | 22.82 | 8.18 |
| | | ABC | 20.93 | 7.50 |
| | | CDGAFS | <u>**19.70**</u> | <u>**7.06**</u> |
| Madelon | 500 | PSO | <u>**74.35**</u> | <u>**14.87**</u> |
| | | ACO | 69.91 | 13.98 |
| | | ABC | 75.12 | 15.02 |
| | | CDGAFS | **75.03** | **15.01** |
| Isolet | 617 | PSO | 141.63 | 22.95 |
| | | ACO | **138.56** | **22.46** |
| | | ABC | 171.73 | 27.83 |
| | | CDGAFS | <u>**135.26**</u> | <u>**21.92**</u> |
| Colon | 2000 | PSO | <u>**12.73**</u> | <u>**0.64**</u> |
| | | ACO | 15.71 | 0.79 |
| | | ABC | 13.22 | 0.66 |
| | | CDGAFS | **12.98** | **0.65** |

Table 5 records the number of selected features of the different wrapper-based feature selection methods for each dataset. It can be observed that, generally, all the methods achieve a significant reduction of dimensionality by selecting only a small portion of the original features. Among the various methods, in the Sonar, Arrhythmia, Madelon, Isolet, and Colon datasets, the proposed method has the best performance among the other methods, selecting only 11.70, 7.20, 13.65, 22.88 and 0.56 %, respectively. Moreover, in the SpamBase dataset, the PSO-base method selected an average of 14.68 % features.

**Table 5:** Average number of selected features of the different wrapper SI-based methods. (Minimum number of selected features is indicated in boldface and underlined and the second best is in boldface)

| Dataset | Number of the original feature | Method | Number of selected features | The ratio of the selected features to the original features (in %) |
|---|---|---|---|---|
| SpamBase | 57 | PSO | 8.98 | 15.75 |
| | | ACO | <u>**8.37**</u> | <u>**14.68**</u> |
| | | ABC | **8.84** | 15.51 |
| | | CDGAFS | 8.93 | 15.67 |
| Sonar | 60 | PSO | 7.34 | 12.23 |
| | | ACO | **7.11** | **11.85** |
| | | ABC | 8.92 | 14.87 |
| | | CDGAFS | <u>**7.02**</u> | <u>**11.70**</u> |
| Arrhythmia | 279 | PSO | **20.41** | 7.32 |
| | | ACO | 21.92 | 7.86 |
| | | ABC | 22.35 | 8.01 |

|  |  | CDGAFS | **20.10** | **7.20** |
|---|---|---|---|---|
| **Madelon** | 500 | PSO | 73.35 | 14.67 |
|  |  | ACO | **69.21** | **13.84** |
|  |  | ABC | 85.43 | 17.09 |
|  |  | CDGAFS | **68.23** | **13.65** |
| **Isolet** | 617 | PSO | **142.53** | 23.10 |
|  |  | ACO | 142.67 | 23.12 |
|  |  | ABC | 164.82 | 26.71 |
|  |  | CDGAFS | **141.18** | **22.88** |
| **Colon** | 2000 | PSO | **11.61** | **0.58** |
|  |  | ACO | 12.79 | 0.64 |
|  |  | ABC | 13.36 | 0.67 |
|  |  | CDGAFS | **11.29** | **0.56** |

Also, several experiments were conducted to compare the execution time of different wrapper EA-based feature selection methods. In these experiments, corresponding execution times (in second) for each method, were reported in Table 6. Due to the fact that the feature selection process and the final classification process are independent, only the execution time for feature selection is reported in the data in this Table. The reported results revealed that the proposed CDGAFS feature selection method has the lowest average execution time overall dataset among all other methods. After the proposed method, PSO-based and ACO-based methods ranked second and third, respectively.

**Table 6:** Average execution time (in second) of wrapper feature selection methods over ten independent runs.

| **Dataset** | **PSO** | **ACO** | **ABC** | **CDGAFS** |
|---|---|---|---|---|
| **SpamBase** | 6.72 | 8.42 | 8.94 | 6.64 |
| **Sonar** | 4.24 | 7.78 | 8.32 | 4.12 |
| **Arrhythmia** | 22.83 | 27.31 | 30.18 | 21.61 |
| **Madelon** | 89.58 | 98.12 | 108.62 | 88.72 |
| **Isolet** | 48.92 | 52.34 | 58.92 | 47.19 |
| **Colon** | 60.82 | 79.12 | 61.41 | 54.95 |
| **Average** | **38.85** | **45.51** | **46.06** | **37.20** |

The performance of CDGAFS for feature selection can be observed in TABLE 3-6; however, the influence of repair operation upon the feature selection process is unclear. Several tests have been conducted to explain exactly how the repair process plays a significant role in CDGAFS for feature selection tasks. Figures 6 and 7 indicate the classification accuracy of GA-based feature selection algorithms in Sonar and SpamBase datasets as well as demonstrate that CDGAFS has been able to find salient features in feature space easily and rapidly. The successful function of CDGAFS repair can be observed clearly in these figures. In these figures, CDGAFS and GAFS denote the GA-based feature selection with proposed repair operation and GA-based feature selection without repair operation, respectively.

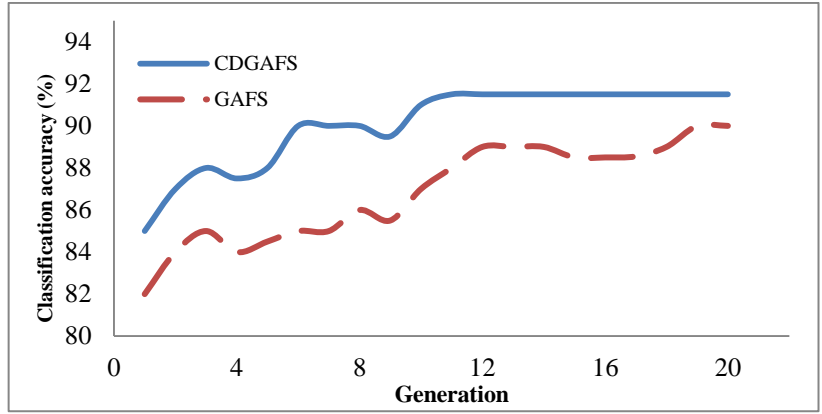

**Fig 6. Comparison of Convergence Process of Proposed Method Using Repair Operator (CDGAFS) and Without Using Repair Operator (GAFS) on SpamBase dataset**

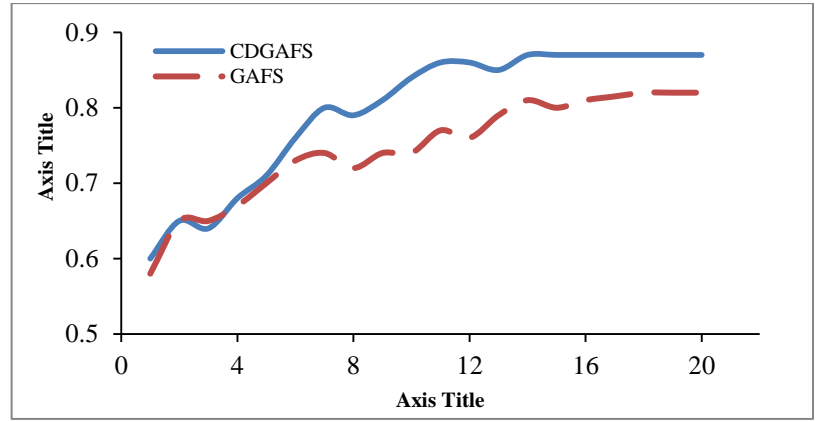

**Fig 7. Comparison of Convergence Process of Proposed Method Using Repair Operator (CDGAFS) and Without Using Repair Operator (GAFS) on Sonar dataset**

## 4.5. Complexity analysis

In this subsection, the computational complexity of the proposed method is calculated. In the first step of the proposed method, the fisher score of all features is measured. The computational complexity of fisher score calculation is $O(ncp)$, where $n$ is the number of the original features and $p$ denotes the number of patterns and $c$ is the number of classes in the dataset.

The first step of the method aims at converting the feature space into a graph and requires $O(n^2 p)$ time steps where $n$ is the number of the original features and $p$ denotes the number of patterns. Moreover, in the next phase, a community detection algorithm is applied to find the feature clusters. The complexity of community detection algorithm is $O(n \log n)$. Then a specific genetic algorithm-based search technique is utilized to choose the final feature set. The search algorithm will be repeated for a number of iterative cycles (i.e., $I$). Thus, the time complexity of this part is $O(IPk f_k)$, where $P$ is the number of the chromosomes in the population, $k$ is the number of the clusters and $f_k$ denotes the time complexity to calculate the fitness function. The time complexity of the KNN classifier is $O(Pn)$. Therefore, the computational complexity of this phase is equal to $O(IP^2 nk)$. Consequently, the final computational complexity of the proposed method is $O(n^2 p + n \log n + IP^2 nk)$, which are reduced to $O(n^2 p + p^2 n)$.

# 5. Conclusions

Feature selection contribute significantly in machine learning and particularly classification task. The computational cost is minimized and the model is designed from simplified data that enhance the overall capabilities of classifiers. A framework was proposed which integrates the advantages of filter and wrapper methods and embeds such a framework into the genetic algorithm in the present article. Some excellent aspects of the proposed technique enhance the efficiencies, the summarization of which is presented as the following. Initially, feature similarities and feature relevance are calculated. Second, CGAFS applies community detection to eliminate redundant features. Hence, the proposed approach picks a certain number of features from each cluster. Also, in this method, unlike previous methods, a multi-objective evolutionary algorithm for the feature selection problem is proposed. The comparison of the performance of the suggested technique with the other feature selection methods is done. The reported results indicate that the proposed method gives higher efficiency, faster convergence, and search efficiency compared to other feature selection methods. For future work, the authors intend to investigate various community detection and social network analysis techniques and apply the maximum clique algorithm for automatically determining the number of clusters and feature clustering.